\documentclass[letterpaper,times]{sae}

\usepackage[dvipsnames]{xcolor}
\definecolor{SAEblue}{RGB}{1,160,233}
\raggedright

\usepackage{cite}
\usepackage[justification=justified, singlelinecheck=false]{caption}  
\DeclareCaptionFont{eightpt}{\fontsize{8pt}{11pt}\selectfont #1}
\usepackage{lineno}
\usepackage[utf8]{inputenc}
\usepackage{changes}
\usepackage{soul}

\usepackage{array}
\newcolumntype{L}[1]{>{\raggedright\let\newline\\\arraybackslash\hspace{0pt}}p{#1}}
\newcolumntype{C}[1]{>{\centering\let\newline\\\arraybackslash\hspace{0pt}}p{#1}}
\newcolumntype{R}[1]{>{\raggedleft\let\newline\\\arraybackslash\hspace{0pt}}p{#1}}

\usepackage{graphicx}
\usepackage{multirow}
\usepackage{amsmath}
\usepackage{csquotes}

\newcommand{\ignore}[1]{}

\usepackage[english]{babel}
\usepackage{amsthm}
\usepackage{amssymb}

\theoremstyle{definition}

\usepackage{hyperref}
\newcommand*{\figref}[2][]{%
  \hyperref[{fig:#2}]{%
    Figure~\ref*{fig:#2}%
    \ifx\\#1\\%
    \else
      \,#1%
    \fi
  }%
}

\usepackage{nameref}
\usepackage{booktabs}

\makeatletter
\def\@seccntformat#1{%
  \expandafter\csname c@#1\endcsname\c@section
  }
\makeatother

\usepackage{subfigure}
\usepackage{multimedia}

\PaperTitle{Real-time Terrain Analysis for Off-road Autonomous Vehicles}
\usepackage{calc} 

\makeatletter 
\renewcommand\@biblabel[1]{#1. } 
\makeatother

\AddAuthor{Edwina Lewis, Aditya Parameshwaran, Laura Redmond, and Yue Wang}{School of Civil and Environmental Engineering and Earth Sciences, Clemson University \\ Department of Mechanical Engineering, Clemson University\\School of Civil and Environmental Engineering and Earth Sciences and Department of Mechanical Engineering, Clemson University \\ Department of Mechanical Engineering, Clemson University}
\PaperNumber{2025-01-8343}
\SAECopyright{2025}

\begin{document}
\maketitle
\section{Abstract}
One challenge for autonomous vehicle (AV) control is the variation in road roughness which can lead to deviations from the intended course or loss of road contact while steering. The aim of this work is to develop a real-time road roughness estimation system using a Bayesian-based calibration routine that takes in axle accelerations from the vehicle and predicts the current road roughness of the terrain. The Bayesian-based calibration method has the advantage of providing posterior distributions and thus giving a quantifiable estimate of the confidence in the prediction that can be used to adjust the control algorithm based on desired risk posture. Within the calibration routine, a Gaussian process model is first used as a surrogate for a simulated half-vehicle model which takes vehicle velocity and road surface roughness (\texttt{GD}) to output the axle acceleration. Then the calibration step takes in the observed axle acceleration and vehicle velocity and calibrates the Gaussian process model to best fit this data. The final result is the posterior distribution of the road surface roughness. To train the Gaussian process model, the half-vehicle model is used to collect vertical axle acceleration data over a range of velocities and road surface roughness levels using Latin Hypercube sampling.  The Bayesian-based calibration method was then implemented in the loop with a Simplex controller to update the velocity limits based on the predicted road surface roughness. To demonstrate the effectiveness of the control algorithm, a stochastically generated surface with different regions of varying road roughness was utilized to test the algorithm's ability to characterize road roughness in real-time, thus showcasing the Simplex control strategy that enhances safety in AV operation. The proposed algorithm has the potential to mitigate the risks associated with road surface roughness, ensuring a safer and more efficient operation for AVs.

\let\thefootnote\relax\footnotetext{DISTRIBUTION STATEMENT A. Approved for public release; distribution is unlimited. OPSEC9325}

\section{Introduction}\label{s2}
A challenge for AVs can arise when the roughness of a road increases. A rougher terrain can lead to issues with vehicle performance and control. There are multiple methods focused on road surface roughness calculation using indirect methods through sensors on or within a vehicle; however, many of the applications in the literature are focused on road maintenance with offline data processing, or real-time characterization of major road defects, rather than real-time road surface roughness characterization needed for AV control. 

One such method of road profile estimation detects anomalies in real-time using jump diffusion process estimation and vehicle velocities \cite{Li2016}. The use of a jump diffusion process estimator and a front half-car model allows for the determination of anomalies that would only affect the left or right side of the car; a pothole would be an anomaly that could be detected with this process. Following a similar trend of many other road surface estimations, the data can be used for maintenance and route determination. In \cite{Li2016}, the authors discuss possibilities of also utilizing the research to create an anomaly map of a region. 

In \cite{Gonzalez2008}, a method of classification of road surface roughness is formulated by using the power spectral density (PSD) of a vehicle's acceleration along a surface to classify the PSD of a road's surface. This ties into the classification of road surface roughness addressed by the International Organization for Standardization (ISO). One thing to note for this method is that it hinges on the knowledge of the transform function associated with vehicle. This method does provide an accurate estimation of the road surface roughness from the acceleration values; however, it doesn't have the emphasis of road surface roughness estimation in real-time that this research project aims to achieve. 

There are even approaches that attempt to determine road surface roughness through sensors not directly attached to the vehicle. An example of this utilizes smartphones as probes to gather large amounts of acceleration data and location data from a vehicle traveling on the roads which lends this method to be more of an an offline approach to classifying the road surface roughness \cite{Douangphachanh2013}. This data can then be utilized to form a relationship between the magnitude of the acceleration and the road surface roughness according to the international roughness index. This method hinges on gathering large amounts of data over the same roads to reduce noise effects from variation in noise levels and phone placement within the vehicles.

There are also multiple cases where a terrain estimation is conducted in real time. One of such cases utilizes a virtual depth image to detect the approaching surface roughness or obstacles \cite{Chang2016}. In this paper, the author aims to determine obstacles and voids so that the robot can avoid them. It uses the virtual depth image to change the boundaries of where the robot can move as it travels. This is useful, but doesn't fully address the intention of an AV control system modifying behavior based on perceived roughness of the terrain. Another case of real time terrain estimation uses a Bayesian inference framework and image-based data to determine the categorical distribution of surface types to map out the elevations of a terrain \cite{Ewen2022}. Through this method, the authors aim to map out the surrounding areas, by yielding a conditional probability of the surface, and determine a probability of traversability in different regions; however, it does not seem that a control algorithm is implemented yet to govern the robots movements through these regions. It should be noted though that the research in \cite{Ewen2022} was performed with a four-legged robot which might not be applicable for the intended use of this paper which is a wheeled AV. 

The prediction of road roughness in this paper is accomplished using a Bayesian inference method. When considering this inference method, there are Bayesian estimation methods utilizing Gaussian process models (GPMs) \cite{Gattiker}, and there are also Bayesian estimation methods utilizing a griddy Gibbs sampler approach \cite{Wright2024}. When considering statistical approaches in general, there are countless more frequentist approaches for calibration. Using a Bayesian estimation method allows for a posterior distribution of calibrated parameters to be gathered, allowing for multiple ways in which the final calibrated parameter can be determined. Some examples include taking the final calibrated parameter as the mean of the distribution, or the final calibrated parameter can be sampled from the distribution itself. Because a posterior distribution of the calibration is gathered, this process allows for real time to decisions to be made while accounting for the uncertainty in the calibrated parameter.

Early research on terrain estimation in the case of mobile wheeled robots typically assumes prior knowledge of terrain roughness, dividing the navigation task into offline and online path planning stages solvable through optimization techniques \cite{doi:10.1177/0278364908097578} \cite{5650821}. In contrast, our approach operates without prior terrain information, computing roughness adaptively during traversal, thereby enabling deployment in previously unknown environments. Other studies employ multi-modal sensor fusion techniques, integrating camera, LiDAR, and 6-DOF IMU sensors to estimate terrain characteristics in advance and subsequently generate safe, optimal trajectories \cite{WeerakoonSLGPM23} \cite{stavens2012self}.We focus on a specific problem dealing with estimating terrain roughness using only the Inertial Measuring Unit (IMU) sensor in the absence of any other sensors. Recent works in outdoor navigation for mobile robots have employed temporal logic \cite{parameshwaran2023safety}, control barrier functions (CBF) \cite{cbf} or reliability based path planners \cite{r2rrt} to measure and maintain safety. For our proposed framework, we use a lightweight Simplex controller, which dynamically modulates the vehicle velocity in real-time thus switching to safer mode of operation. The Simplex architecture allows a high-performance primary controller to be complimented with a high-assurance safety controller \cite{seto1998simplex}.

Each approach has their strengths and drawbacks. For this research paper, a Bayesian inference method that utilizes a GPM as a surrogate for the the vehicle-terrain model will be used. The method used in this paper will have certain advantages compared to the other reviewed papers. This method will deal with the classification of a surface with a singular numeric parameter which will be calibrated. From this, a posterior distribution of the calibrated parameter is formulated, allowing for a real time risk characterization in response to road roughness. This method would not require any prior knowledge on the terrain, and it doesn't use an offline approach to the terrain roughness estimation. It should be noted that this method is intended for use on surfaces without large voids or obstacles impeding movement.

\begin{figure}[!ht]
    \centering
    \includegraphics[scale = 0.5,width = 90mm]{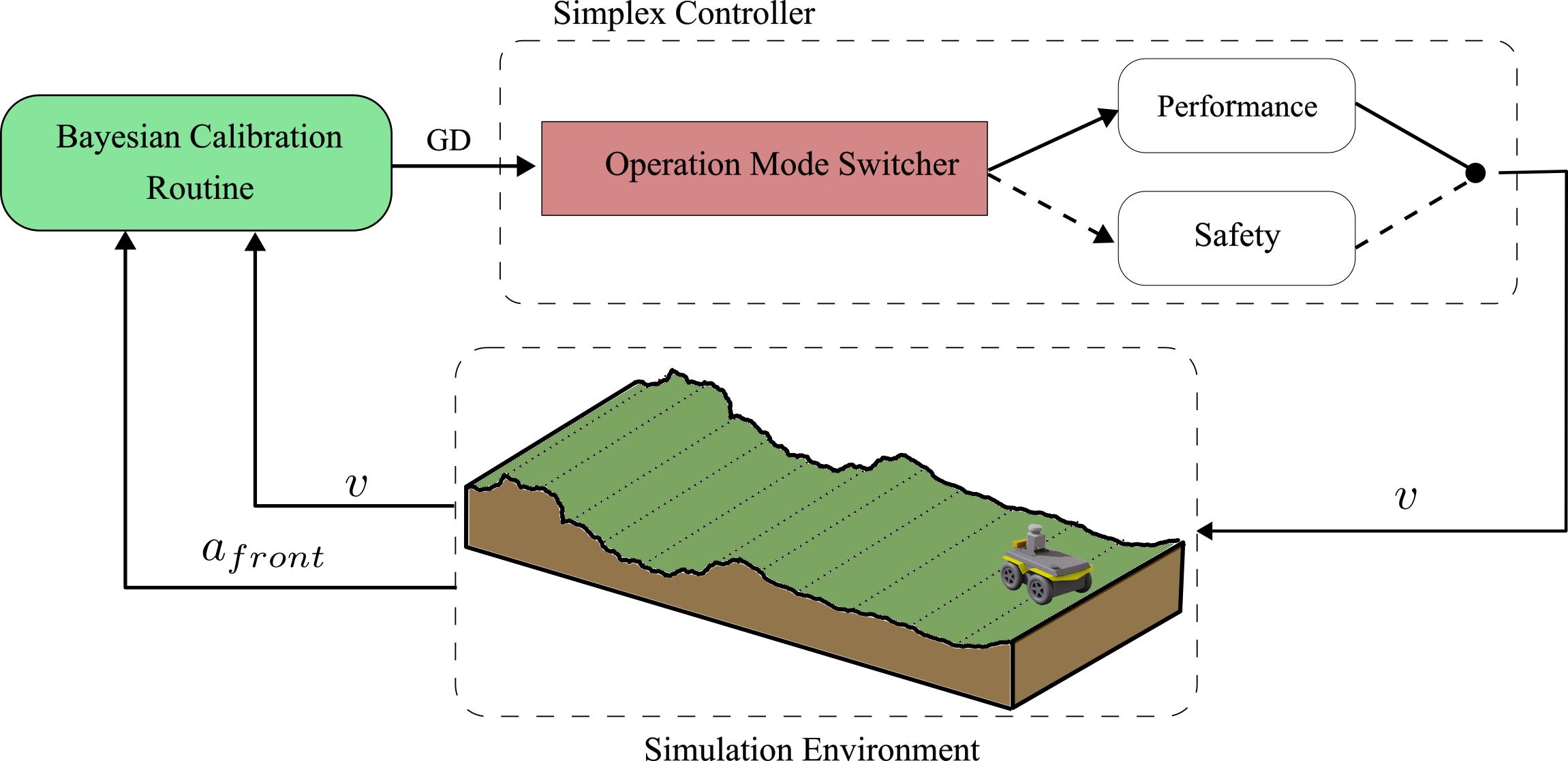}
    \caption{The trained Bayesian Calibration Routine (BCR) utilizes the variance of the longitudinal acceleration $(\boldsymbol{a_{front}})$ and linear velocity $(v)$ from the simulated robot as input parameters to estimate the terrain roughness, quantified by the \texttt{GD} value. Subsequently, the \texttt{GD} value is employed to switch the operational mode of the Simplex controller, thereby modulating the control linear velocity $v$ of the simulated robot. This facilitates real-time feedback control of the robot's velocity in response to varying terrain roughness. }
    \label{fig:approach}
\end{figure}

In this paper, a process for updating the AV speed based on predicted road surface using only data from the IMU is developed utilizing a GPM-based Bayesian calibration routine paired with a Simplex controller. The goal is to have this routine act as a continuous closed loop system that gives constant feedback to the controller system of the AV. This demonstrates the potential to use such algorithms for safety control of a vehicle by regulating speed on \enquote{rougher} roads. To start this routine, a calibration metric must be chosen that is sensitive to both the known and unknown parameters of the model. Next, a Bayesian-based calibration method will be utilized to determine the road surface roughness of the path that the vehicle is traveling on. Once, the road surface roughness is determined, then a method of controlling the AV must be chosen where it can make a reasonable safety change based on the roughness of the road. Though the vehicle in this project is at a smaller scale, the changes implemented here would be representative of safety changes that are implemented to help prevent course deviation and loss of road contact while the vehicle is operating. An overview of the approach can be seen in Figure \ref{fig:approach}.

The outline of the paper is as follows. First, the preliminaries and background to conduct the simulations are discussed in \nameref{Preliminaries}. Next, the GPM-based Bayesian calibration method for determination of the road surface roughness is explained in \nameref{Analysis Methodology}. Additionally, the details of the controls in both the autonomy of the vehicle are outlined in \nameref{Analysis Methodology}. The set up and results of running the simulation to display the proposed solution is discussed in \nameref{Experiments}.Finally, further implications of the experiments will be discussed in the \nameref{Conclusion} section.
 \color{black}

\section{Vehicle-Terrain Simulations}\label{Preliminaries}

\subsection{Half-Vehicle Model} \label{HVM}
\begin{figure}[!ht]
    \centering
    \includegraphics[scale = 0.5,width = 90mm]{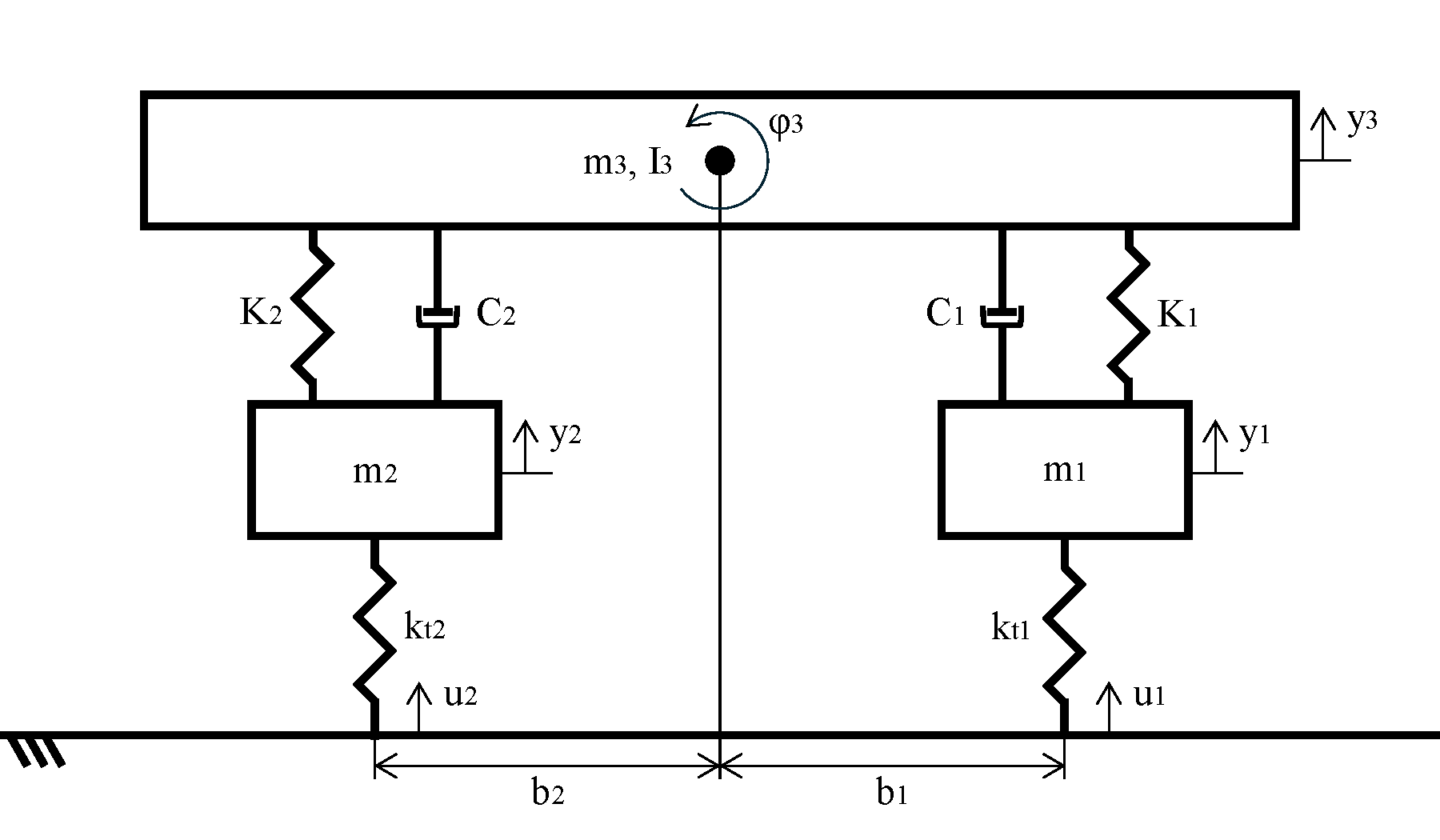}
    \caption{Half-Vehicle Model}
    \label{fig:half_car_model}
\end{figure}

The dynamic behavior of the vehicle in response to road-induced excitations was evaluated using a longitudinal half-car model \cite{half-vehicle}, illustrated in Fig. \ref{fig:half_car_model}. This model includes the unsprung masses of the front and rear wheels, denoted as $m_1$ and $m_2$, respectively, as well as the sprung mass of the vehicle body, represented by $m_3$. The front and rear suspensions, which consist of linear springs and viscous dampers, connect the vehicle body and wheels. These suspension elements are described by the stiffness constants $K_1$ and $K_2$ and the damping coefficients $C_1$ and $C_2$. The tires’ elastic behavior is modeled by springs with stiffness values $k_{t1}$ and $k_{t2}$. The longitudinal half-car model possesses four degrees of freedom: vertical movements of the masses ($y_1$, $y_2$, $y_3$) and the rotational motion of the vehicle body about its center of mass, commonly referred to as pitch ($\varphi_3$).

The equations governing the dynamics of the half-vehicle model are shown below:
\begin{equation}
\begin{aligned}
    &m_1 \ddot{y}_1 + C_1\left(-\dot{y}_1 + \dot{y}_3 + b_1 \dot{\varphi}_3\right) + K_1\left(-y_1 + y_3 + b_1 \varphi_3\right)\\ 
    &+ k_{t 1}\left(y_1 - u_1\right) = 0 \\
    &m_2 \ddot{y}_2 + C_2\left(-\dot{y}_2 + \dot{y}_3 - b_2 \dot{\varphi}_3\right) + K_2\left(-y_2 + y_3 - b_2 \varphi_3\right)\\
    &+ k_{t 2}\left(y_2 - u_2\right) = 0 \\
    &m_3 \ddot{y}_3 + C_1\left(-\dot{y}_1 + \dot{y}_3 + b_1 \dot{\varphi}_3\right) + C_2\left(-\dot{y}_2 + \dot{y}_3 - b_2 \dot{\varphi}_3\right) \\
    &\qquad+ K_1\left(y_3 - y_1 + b_1 \varphi_3\right) + K_2\left(-y_2 + y_3 - b_2 \varphi_3\right) = 0 \\
    &I_3 \ddot{\varphi}_3 + C_1 b_1\left(-\dot{y}_1 + \dot{y}_3 + b_1 \dot{\varphi}_3\right) + C_2 b_2\left(-\dot{y}_2 + \dot{y}_3 - b_2 \dot{\varphi}_3\right) \\
    &\qquad+ K_1 b_1\left(y_3 - y_1 + b_1 \varphi_3\right) + K_2 b_2\left(-y_2 + y_3 - b_2 \varphi_3\right) = 0
\end{aligned}
\end{equation}

where $y$, $\dot{y}$, $\ddot{y}$, $\dot{\varphi}$, and $\ddot{\varphi}$ represent the vertical displacement, vertical velocity, vertical acceleration, angular velocity, and angular acceleration, respectively \cite{Frankovsky2011,Delyova2012}. The parameters $b_1$ and $b_2$ denote the distances from the vehicle's center to the front and rear suspensions, respectively.

\subsection{Road Surface Roughness Classifications}
There are multiple ways to classify road surface roughness; however, this project needed a road classification method that was quantifiable. This lead to the use of the definition of road surface roughness according to the ISO 8608 standard \cite{ISO8608}. This standard characterizes the road surface roughness by the fitted smooth power spectral density (PSD) of the vertical displacement of a surface. From this standard, this PSD of displacements is referred to as the function \texttt{GD}(n) where n is the reference spatial frequency of the value. For simplicity of this model, the reference spatial frequency is chosen to be 0.1 cycles/m to align with the level of service references found in the standard \cite{ISO8608}. In further references, the term \texttt{GD} will refer to the value of the fitted smooth PSD of displacements at the spatial frequency of 0.1 cycles/m multiplied by a factor of 10\textsuperscript{6}. For clarity, \texttt{GD} is defined by the following relationship:
\begin{equation} \label{GD}
    \begin{aligned}
    \texttt{GD} = \texttt{GD}(0.1 \frac{cycles}{m}) * 10^6
    \end{aligned}
\end{equation}
As alluded to, the ISO 8608 standard provides a list of general \texttt{GD} values that correspond to different road classes from A to H. A road class of A corresponds to a smooth road, and every road class after that corresponds to an increasingly rougher road. Table \ref{table:1} shows how the upper and lower limits of \texttt{GD} for corresponding levels of service \cite{ISO8608}. 

\begin{table}[h!]
\begin{center}
    \caption{Road roughness levels corresponding to \texttt{GD} quantification \cite{ISO8608}.}
    \label{table:1}
    \begin{tabular} {c|c c}
    Road Class &Lower Limit (\texttt{GD})& Upper Limit (\texttt{GD}) \\
    \hline
    A&-&32 \\
    B&32& 28 \\
    C&128&512 \\
    D& 512&2,048 \\
    E& 2,048&8,192 \\
    F& 8,192&32,768 \\
    G& 32,768&131,072 \\
    H& 131,072&- \\
    \end{tabular}
\end{center}
\end{table}

In this project, the effect of \texttt{GD} was seen by having it as an input parameter when creating a stochastic road profile whose PSD of vertical displacements had the wanted \texttt{GD} value at a spatial frequency of 0.1 cycles/m. To create this stochastic profile, a code was utilized that was originally constructed to simulate a road profile for a structural health monitoring project \cite{Locke2022}. This code generates a road profile when given a length of terrain, the distance between each defined point, and a prescribed \texttt{GD} value for the surface. Typically the distance between the defined path was set as the speed at which the vehicle was running divided by the expected sampling frequency. Because a road profile wasn't able to be defined at an infinitely small length along the road, the profile had to be interpolated such that a height was defined every 6 mm. 

\begin{figure}[!ht]
    \centering
    \includegraphics[scale = 1.0,width = 90mm]{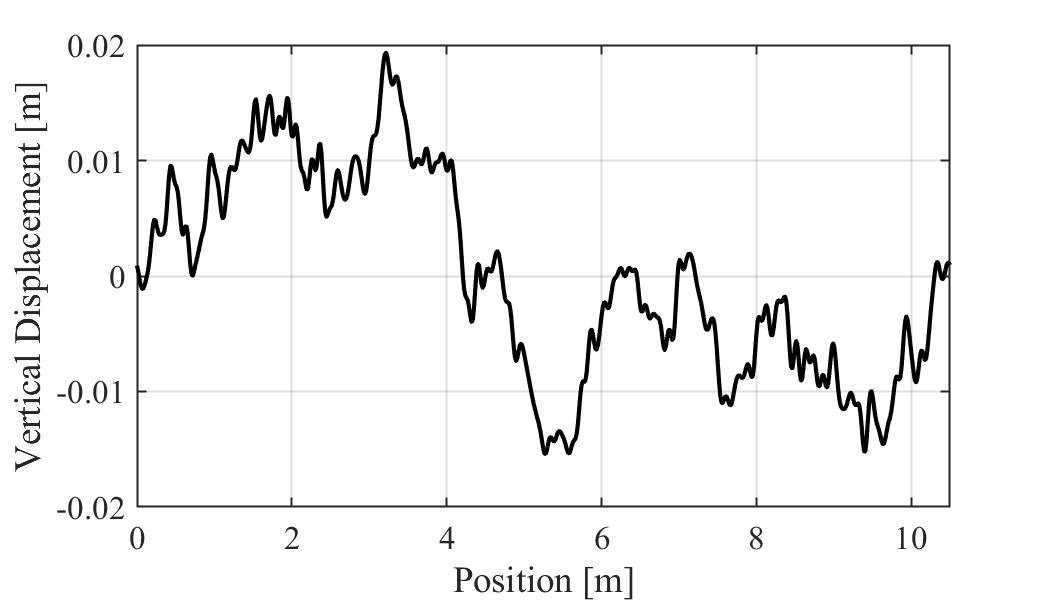}
    \caption{The vertical displacement of an example profile with a prescribed \texttt{GD} of 450}
    \label{fig:GD450_vert_disp}
\end{figure}

\begin{figure}[!ht]
    \centering
    \includegraphics[scale = 1.0,width = 90mm]{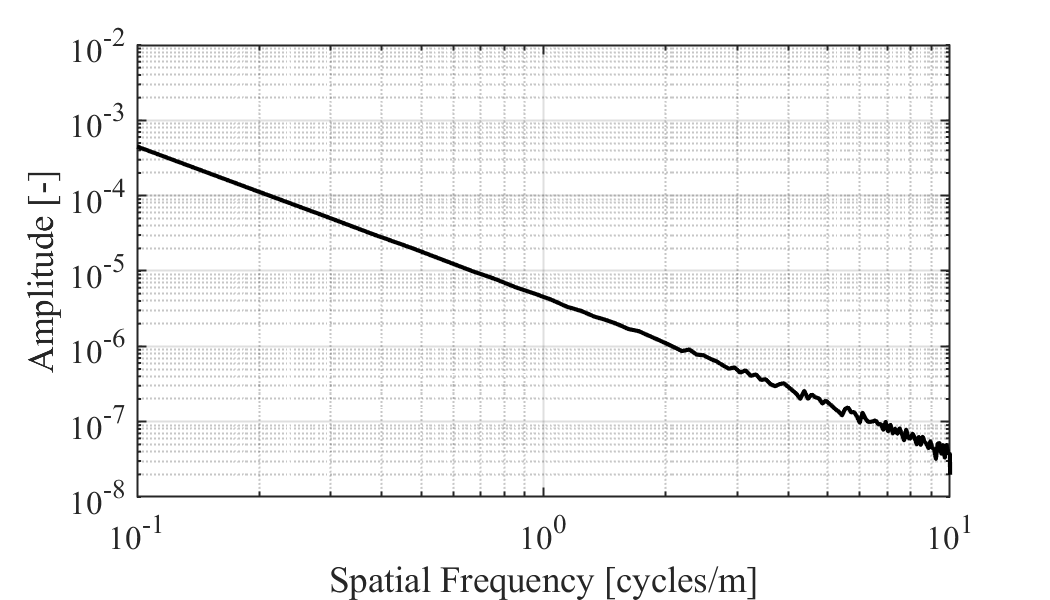}
    \caption{The power spectral density of the vertical displacements shown in Fig. \ref{fig:GD450_vert_disp}.}
    \label{fig:GD450_PSD}
\end{figure}

To ensure that the road profile still corresponded to the proper \texttt{GD}, a surface was generated with a prescribed \texttt{GD} value of 450. The vertical displacement of this generated surface can be seen in Fig. \ref{fig:GD450_vert_disp}. Then, the fast Fourier transform of these vertical displacements was utilized to compare the PSD at a spatial frequency of 0.1 cycles/m to the prescribed \texttt{GD} value.  As seen in Fig. \ref{fig:GD450_PSD}, at a spatial frequency of 0.1 cycles/m, the PSD has a value of 449 which does closely correspond to the prescribed \texttt{GD} value of 450.

\section{Real-time Terrain Analysis Methodology}\label{Analysis Methodology}

\subsection{Bayesian Calibration Routine (BCR)}\label{BCR}
The proposed solution for the stated problem is to utilize a GPM-based Bayesian Calibration Routine that implements simplex control method to give safety feedback to the AV. The following section will outline the mechanics of the routine which utilizes, at its core, a GPM, Markov-chain Monte Carlo (MCMC) method Bayesian-based code which is adapted from one developed by Los Alamos National Laboratory \cite{GPMSA_Code}. 

Before discussing Bayesian calibration methods, Bayes' theorem is introduced. Bayes' rule can be derived from the definition of a joint probability \cite{Gelman2013} statement and described as:
\begin{equation} \label{eq:Bayes}
    \begin{aligned}
    p(\boldsymbol{\theta}|\boldsymbol{f})\propto p(\boldsymbol{\theta})p(\boldsymbol{f}|\boldsymbol{\theta})
    \end{aligned}
\end{equation}
In the context of a model, the random variable $\boldsymbol{\theta}$ represents the vector unknown parameters of a data model, and the random variable $\boldsymbol{f}$ represents the vector of observed results from a data model which utilizes $\boldsymbol{\theta}$ as an input. This allows for Bayes' rule to give the posterior distribution of $\boldsymbol{\theta}$ if the prior, $p(\boldsymbol{\theta})$, and the conditional, $p(\boldsymbol{f}|\boldsymbol{\theta})$, distributions are known. The prior distribution $p(\boldsymbol{\theta})$ is determined to represent knowledge that might exist about the unknown parameter. To obtain $p(\boldsymbol{f}|\boldsymbol{\theta})$, a given data model is run a specified amount of times with the $\boldsymbol{\theta}$ values drawn from $p(\boldsymbol{\theta})$. In terms of model calibration, the parameter $\boldsymbol{x}$ can also be utilized to represent the vector of known parameters of the data model. This means that when in the calibration environment, the following version of Bayes' rule is valid:
\begin{equation} \label{eq2}
    \begin{aligned}
    p(\boldsymbol{\theta}|\boldsymbol{f})\propto p(\boldsymbol{\theta})p(\boldsymbol{f}|\boldsymbol{\theta},\boldsymbol{x})
    \end{aligned}
\end{equation}
For the context of this model, $\boldsymbol{f}$, $\boldsymbol{x}$, and $\boldsymbol{\theta}$ are 1 dimensional. From here on, in regards to the Bayesian calibration, the nomenclature will be that $f$ is the observation from the data model; $v$ is the known parameter of the data model, velocity (replacing $\boldsymbol{x}$); and, \texttt{GD} is the unknown parameter of the data model, the quantification of road roughness (replacing $\boldsymbol{\theta}$).

For this project, the $v$ parameter represents the velocities of the vehicle on the domain of 0.5 m/s to 2 m/s, and the \texttt{GD} parameter represents the road roughness values on the domain of 200 to 600. The domain of $v$ was chosen according to the physical limits of the vehicle. The vehicle has a maximum speed of 2 m/s, and the lower bound was chosen to be 0.5 m/s as the lowest practical speed.  The domain of \texttt{GD} was chosen when considering the feasibility of obtaining simulated observations to construct the conditional probability of the data model $p(f|{\texttt{GD}},v)$; this domain was from 200 to 600 which falls in the B and C road classes \cite{ISO8608}.

\begin{figure}[!ht]
    \centering
    \includegraphics[scale = 0.5, width = 90mm]{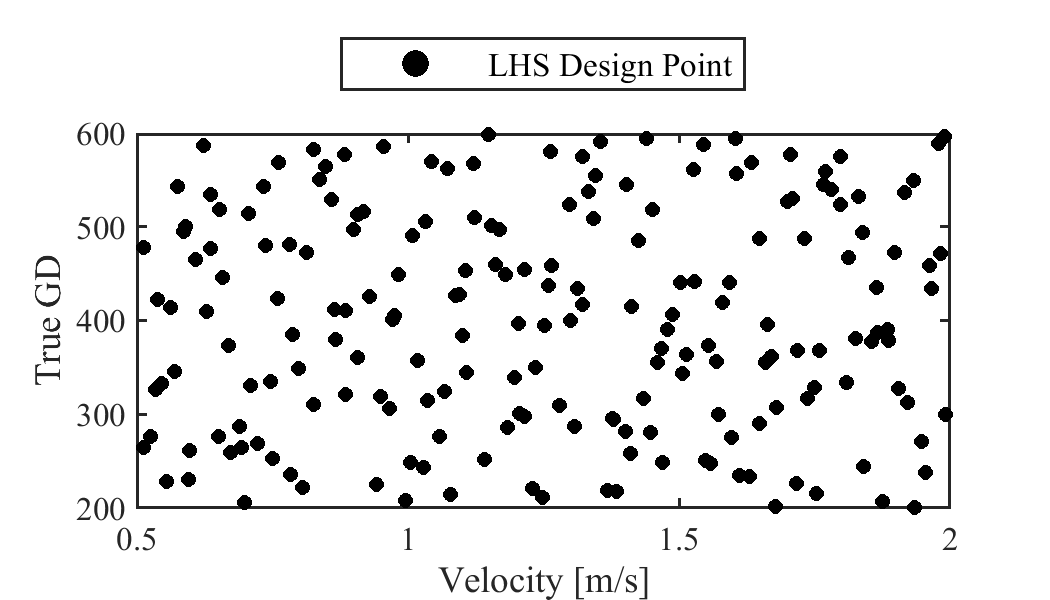}
    \caption{The locations of each of the design points used to train the surrogate model that is used to calibrate in the BCR.}
    \label{LHS_design}
\end{figure}

The \texttt{GD} parameter was assumed to have a uniform distribution over its domain. It is described as the following distribution:
 \begin{equation} \label{eq3}
    \begin{aligned}
    p({\texttt{GD}}) = \left\{\begin{array}{cc}
    \frac{1}{600-200} & 200\leq{\texttt{GD}}\leq 600 \\
    \\
    0 & (-\infty, 200) \ \cup \ (600, +\infty)
    \end{array}\right.
    \end{aligned}
\end{equation}
Similarly, it was assumed that the $v$ parameter was assumed to have a uniform distribution over its domain. It is described as the following distribution:
 \begin{equation} \label{eq4}
    \begin{aligned}
    p(v) = \left\{\begin{array}{cc}
    \frac{1}{2.0 \frac{m}{s}-0.5 \frac{m}{s}} & 0.5 \frac{m}{s}\leq v\leq 2.0 \frac{m}{s} \\
    \\
    0 & (-\infty,  0.5 \frac{m}{s}) \ \cup (\ 2.0 \frac{m}{s}, +\infty)
    \end{array}\right.
    \end{aligned}
\end{equation}
To create the surrogate model, the half-vehicle model needs to be run at a combination of $v$ and \texttt{GD} values over a defined space. Rather than directly running the simulation at a dense grid of  $v$ and \texttt{GD} settings, a Latin Hypercube Sampling (LHS) algorithm \cite{MatlabLHS} and the distributions described in Eqns. \ref{eq3} and \ref{eq4}were utilized to achieve a space-filling design onsisting of 198 points which would the be used to train a GPM. The specifics of the LHS design are described by McKay, and this paper utilizes the LHS design in a one-dimensional approach \cite{McKayLHS}. Fig. \ref{LHS_design} shows how the LHS design is able to fill the space. Though it is space-filling, there are still gaps that were unavoidable unless a very large number of points is used. If computation and time constraints are not a concern, a very dense grid can be used for training the surrogate model.

The code originating from Los Alamos \cite{GPMSA_Code} focuses on creating a surrogate emulator model and then predicts what the unknown parameter is based on observations. The overall process of this code is described in \cite{Gattiker} where the code first utilizes the simulated observations and experimental observations to create a surrogate model, and then next, it utilizes MCMCs to sample the model and obtain the parameters of the model. Once the surrogate model is created, it can then be used to calibrate the unknown parameters of the model. To get this process close to real time, a version of the code was utilized that migrated the process from \texttt{Matlab} to \texttt{Julia}; additionally, this code was restructured so that the surrogate model generation was able to be decoupled from the experimental observations \cite{Wright2024}. This feature was key to allow the surrogate model to be constructed a priori and have it such that the BCR only needs to call the parameters of the surrogate model when in the routine.

When conducting the routine in \texttt{Julia}, the amount of iterations and number of Markov-chain Monte Carlo (MCMC) simulations must be modified when creating the surrogate model and running the calibration. For this specific research project the number of iterations for the surrogate model was chosen to be 30,000. When looking at the portion of the calibration done actively in the BCR, the number of MCMC simulations was chosen to be 5000 with the first 1000 set to be burned. Also within the active calibration, a step size needs to be determined when choosing the next parameters for the next iteration. This can be controlled by setting the number of iterations for the step size algorithm and the amount to burn at each step which was set to be 1000 and 50, respectively. There are many other parameters that affect various aspects of the specific calibration, but these were the main parameters that were adjusted to achieve better performance of the BCR.

The implemented terrain analysis routine begins by generating a moving buffer of the last 1000 points of acceleration data from the vehicle running over a surface and creating the $f$ observation from such data. The observed $f$, is then used to calibrate the GPM and generate the posterior \texttt{GD} distribution. From which, the calibrated \texttt{GD} is taken to be the average of the \texttt{GD} values. Once the BCR determines the \texttt{GD} value of the surface, the Simplex controller will determine the mode of operation, thus activating the appropriate control module. Depending on the state of the controller, the linear velocity of the AV is modulated to drive it safely on the terrain. At the end of this routine, the entire process starts over, and acceleration data is gathered by collecting the last 1000 points of acceleration data in the moving buffer. Because the buffer size is not directly related to the length the AV travels while the routine runs, there are cases of overlapping. 

\subsubsection{Calibration Metrics}\label{calibration_metrics}
In order to perform the Bayesian-based calibration routine, an observation output of the data model, $f$, must be chosen. The goal for $f$ is to be an efficient metric that can be utilized to calibrate future data, which means it needs to be sensitive to the \texttt{GD} and $v$ parameters, which act as inputs to the data model by utilizing \texttt{GD} and $v$ to construct the surfaces for the vehicle to traverse. To construct the dynamic response of the vehicle, $f$, we chose the half-vehicle model as described in the \nameref{HVM} subsection with a few changes based on the robot used in our simulations. The half-vehicle model for our simulation is that of the Clearpath Jackal, which is a rigid-axle robot. Hence, the damping coefficients are negligible $(C_1,C_2) \approx 0$, and the spring stiffnesses are replaced to be extremely stiff $(K_1,K_2) \approx \infty$.

In other cases of similar types of calibrations \cite{Li2016,Gonzalez2008}, acceleration data of the vehicle was chosen to be the main source of output for vehicle behavior, so to be consistent, this research project decided to utilize the acceleration response as well. For specificity, the vertical acceleration data at the axle location of the vehicle was recorded, and this data set will be referred to as $a_{front}$. It should be noted that $a_{front}$ is a vectors of length $T$ where $T$ is the number of time steps for the vehicle to traverse the entirety of its prescribed profile. And, as follows, $a_{front, t}$ refers to the acceleration at a specific time step $t$. 

The variance of $\boldsymbol{a_{front}}$, normalized with respect to $T$ rather than $T - 1$ was chosen as an efficient parameter for calibration. It follows the relationship below:
\begin{equation} \label{eq5}
    \begin{aligned}
    f = 
    \frac{1}{T}*\sum_{t=1}^{T} (a_{front, t} - \mu)^2
    \end{aligned}
\end{equation}
As stated within the discussion of the routine, the distribution of the $f$ observations is representative of the conditional distribution $p(f|\theta,v)$. 

\subsubsection{Assessment of the GPM-based Bayesian Calibration Method}

To determine the adequacy of the calibration metric in an idealized, no noise environment, a post predictive check was done over the domain of the model. This was done by setting a defined mesh with differing $v$ and $\theta$ values of points that were not originally existing in the simulated observations. This allows for a general look at the fit of the model across the entire domain. For the given domain of velocities from 0.5 m/s to 2 m/s and \texttt{GD} from 200 to 600, 12 points were chosen across the domain to analyze the accuracy of the \texttt{GD} predictions. The points along the domain for $v$ that were chosen were the velocities 0.75 m/s, 1.25 m/s, and 1.75 m/s. The points along the domain for $\theta$ that were chosen were the \texttt{GD} values of 250, 300, 350, 400, 450, 500, and 550. The following table shows the percent error on the mean \texttt{GD} from the posterior distribution of calibration compared to the true \texttt{GD} used to generate the simulated data for calibration. The table is organized by the cases examined where the rows correspond to the true \texttt{GD} and the columns correspond the the velocity. As seen here, when just purely looking at the mean of the reported $\theta$ values, the calibration isn't far off, and accuracy is worst near the scenarios near the edge of the training domain, particularly for lower \texttt{GD} values and low velocity.

\begin{table}[!ht]
\begin{center}
    \caption{Percent error check between calibrated \texttt{GD} and prescribed \texttt{GD}.}
    \label{table:2}
    \begin{tabular} {c c | c c c }
    & &\multicolumn{3}{c}{Prescribed Velocity (m/s)}\\
    & &0.75&1.25&1.75\\
    \hline
    \multirow{7}{*}{Prescribed \texttt{GD}}
    &250&78.2\%&12.3\%&46.8\% \\
    &300&33.6\%&23.6\%&7.2\%\\
    &350&24.2\%&0.2\% &14.2\%\\
    &400&2.2\%&4.0\%&30.9\%\\
    &450&4.9\%&16.1\%&1.5\% \\
    &500&18.5\%&3.1\%&5.7\%\\
    &550&22.1\%&7.4\%&19.2\% \\
    \end{tabular}
\end{center}
\end{table}

In addition to looking at the mean calibrated \texttt{GD} value, the posterior distribution of $\theta$ should be examined. With the type of Bayesian calibration methods that are used, the expected posterior distribution of $\theta$ should be unimodal and approximately normal. As seen in Fig. \ref{fig4} located in \nameref{appendix_A}, most of the cases where there is a low percent error, there is also a unimodal posterior distribution of $\theta$. It should be noted that this posterior distribution analysis was done for all of the checked prescribed \texttt{GD} values seen in Table \ref{table:2}, and the histograms of which are found in \nameref{appendix_A}, at Fig. \ref{fig4}. Generally, it is seen that at a velocity of 0.75 m/s we have a poor posterior distribution, and thus, we can't say that we have an accurate calibration. As corroborated in Table \ref{table:2}, Fig. \ref{fig4} shows that as the prescribed \texttt{GD} approaches the boundaries, it has trouble with providing a good calibration that displays the approximate normalcy of the posterior $\theta$ distribution. Due to the stochastic nature of the surface generation and the nature of LHS, there may also be cases where the calibrator won't accurately calibrate the roughness of the surface. This can be seen in Fig. \ref{fig4} with the case with a prescribed \texttt{GD} of 400 and $v$ of 1.75 m/s and the case with a prescribed \texttt{GD} of 250 and $v$ of 1.75 m/s.

Utilizing the knowledge gathered from Table \ref{table:2} and Figure \ref{fig4}, it was decided that the calibration routine doesn't behave as desired for a velocity at or below 0.75 m/s. And, it overall behaves acceptably in the other cases tested except, caution should be used when nearing the boundaries of the domains for $\theta$ and $v$. 
In the test cases for this paper, the two extremes of the testing scenario are at a case where the prescribed \texttt{GD} value is 500 at a speed of 1.0 m/s and at a case where the prescribed \texttt{GD} value is 300 at a speed of 2.0 m/s. Though these cases are near the boundaries of the domain, they were seen as acceptable since there weren't high percent errors in the posterior distributions near these cases, as seen in Table \ref{table:2}. If higher accuracy were required over a larger domain the GPM could be retrained with additional simulations to expand both the domain and the density of the simulation cases used to inform the GPM.

\subsection{Roughness based Simplex Control}
In the Simplex control architecture, a high-performance control module is complemented by a robust safety controller. Should a fault occur in the performance controller, the system can seamlessly transition to the safety controller, which prioritizes reliability over performance. In our implementation, the performance module of the Simplex architecture is a proportional controller that adjusts the robot's linear velocity based on the \texttt{GD} value. Conversely, the safety module will maintain a fixed linear velocity of 1.0 m/s to ensure consistent and safe operation. The switch between the 2 control modules occurs when the BCR measures a runtime \texttt{GD} value higher than the threshold value ($\texttt{GD}_{max}$). The goal of the Simplex controller is to switch between the performance and safety control modules such that the AV travels for better performance while guaranteeing safety. The controller can be mathematically defined as follows:
\begin{equation}
v(t) = 
\begin{cases} 
K_p (\frac{(\texttt{GD}_{min} + \texttt{GD}_{max})}{2}- \texttt{GD}(t)), & \text{if }  \texttt{GD}(t) \in [\texttt{GD}_{min},\texttt{GD}_{max}] \\
1.0 \, \text{m/s}, & \text{if } \texttt{GD}(t) > \texttt{GD}_{max}
\end{cases}\label{control_eq}
\end{equation}
where $v(t)$ is the linear velocity the control input to the AV at time $t$ and $K_p \in \mathbb{R}^+$ is the proportional gain. The range of GD values in which the system is said to be in performance mode is given by $[\texttt{GD}_{min},\texttt{GD}_{max}]$. For this paper, $\texttt{GD}_{min}$ is set to be 250, and $\texttt{GD}_{max}$ is set to be 350. \texttt{GD(t)} is the terrain roughness value calculated by the BCR model during run-time. The maximum allowable velocity of the AV is 2.0 m/s. Later on in the \nameref{Experiments} section we will show the performance of this control strategy on a mobile robot traversing rough terrains with varying \texttt{GD} values.

\section{Simulated Experiments}\label{Experiments}
In this section, we present the data collection method and the results to evaluate the efficacy of our proposed framework. Prior to running the BCR for the experiment, the \texttt{Matlab} and \texttt{Julia} environments were utilized to create the data model that is used to run the routine. 

To train the BCR and test the overall framework, we designed terrains using \texttt{Blender} software. The simulations were run on \texttt{NVIDIA's Isaac Sim} environment, alongside a \texttt{ROS2} and \texttt{C++} framework. The robot used in the simulations was a Clearpath Jackal robot, selected from the \texttt{Isaac Sim} assets library. The overall framework loop was run on an Alienware desktop with a 3.8GHz 8-core CPU and 64 GB of RAM, running Linux OS Version 20.04 LTS.

\subsection{Simulations}
To generate the conditional distribution $p(f|{\texttt{GD}},v)$ for the BCR, the \texttt{Isaac Sim} environment was utilized. Profiles with differing $v$ and $\theta$ parameters were generated for the AV to traverse the terrain, and the output variable $f$ was then obtained. A total of 198 profiles were generated using a Latin Hypercube Sampling (LHS) procedure. The parameters $v$ and $\theta$ were sampled according to the prior distributions defined in Equations \ref{eq4} and \ref{eq3}, respectively. The domain limits were set to $0.5 \leq v \leq 2.0 \ \text{m/s}$ and $200 \leq \texttt{GD} \leq 600$ based on the considerations described in the \nameref{BCR} subsection. There were two different surrogate models trained. The first surrogate model assumes there was no measurement noise in collection. The second surrogate model assumes there is measurement noise, $\epsilon$, in collection. This addition of measurement noise is simulated as having a normal distribution centering around zero with a standard deviation $\sigma$ (normalized to the number of measurement points minus 1) of 1. The distribution of this noise is shown in Eq. \ref{meas_eq}. 

\begin{equation}
\label{meas_eq}
    \epsilon \sim \text{N}(0, \sigma = 1)
\end{equation}

\begin{figure}[ht]
    \centering
    \includegraphics[scale=0.5, width=90mm]{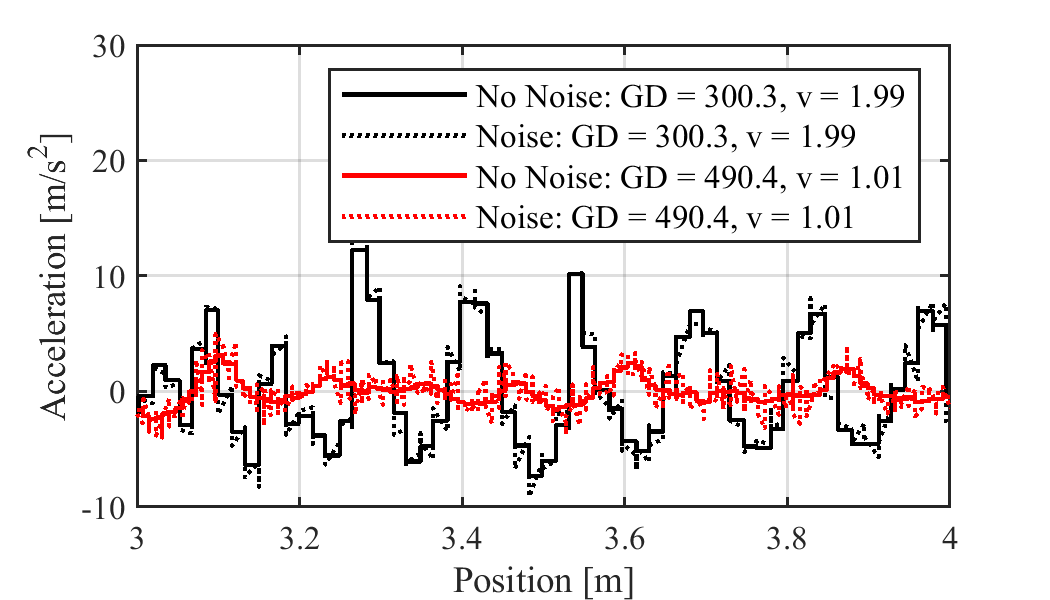}
    \caption{A short comparison of the relative accelerations seen between two cases where both cases show their values with and without noise (the sampling rate for both cases are the same, but a higher velocity leads to less points over the same length of surface).}
    \label{noise_comp}
\end{figure}

The level of noise was chosen such that the signal to noise ratio is reasonable, or as close to reasonable over the domain of the procedure. The general distribution of the signal to noise ratio that would be found in the surrogate model training and data and test data was calculated by generating the noise signals by sampling from \ref{meas_eq} at each measurement point. Then, the signal power of the acceleration responses and the generated noise was calculated to determine the signal to noise ratio. The chosen distribution of error results in a lower signal to noise ratio occurs with lower velocities. The level of error does not change with velocity, and the measurement recorded, acceleration, increases as the velocity of the vehicle increases. This leads to a situation where the noise variation is the same, but the magnitude of the recording increases as velocity increases. An example of how a signal looks with and without noise is shown in Figure \ref{noise_comp}. Additionally, this figure shows the relative magnitudes of acceleration that can exist within the domain. It should be noted that due to noise being applied from random sampling, the signal to noise ratios will vary each time the measurement is sampled.

\begin{figure}[h]
    \centering
    \includegraphics[scale = 0.3,width = 90mm]{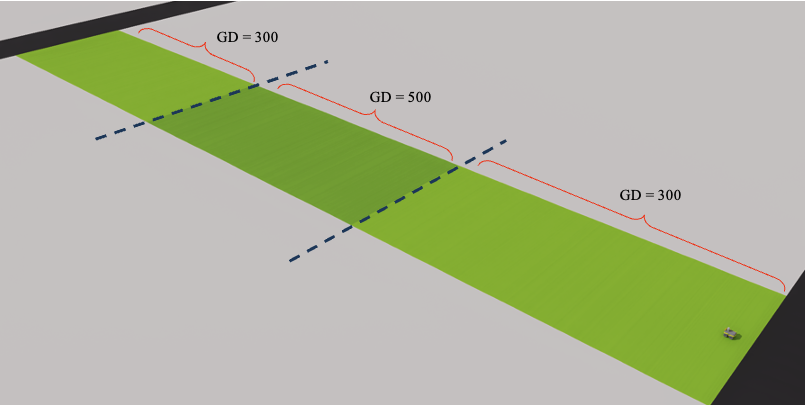}
    \caption{Depiction of the experimental simulation in the NVIDIA \texttt{Isaac Sim} environment. The \textcolor{blue}{blue} line depicts the separation of the terrains with different \texttt{GD} values.}
    \label{fig:simulation}
\end{figure}

\begin{figure}[h]
    \centering
    \includegraphics[scale = 0.5,width = 90mm]{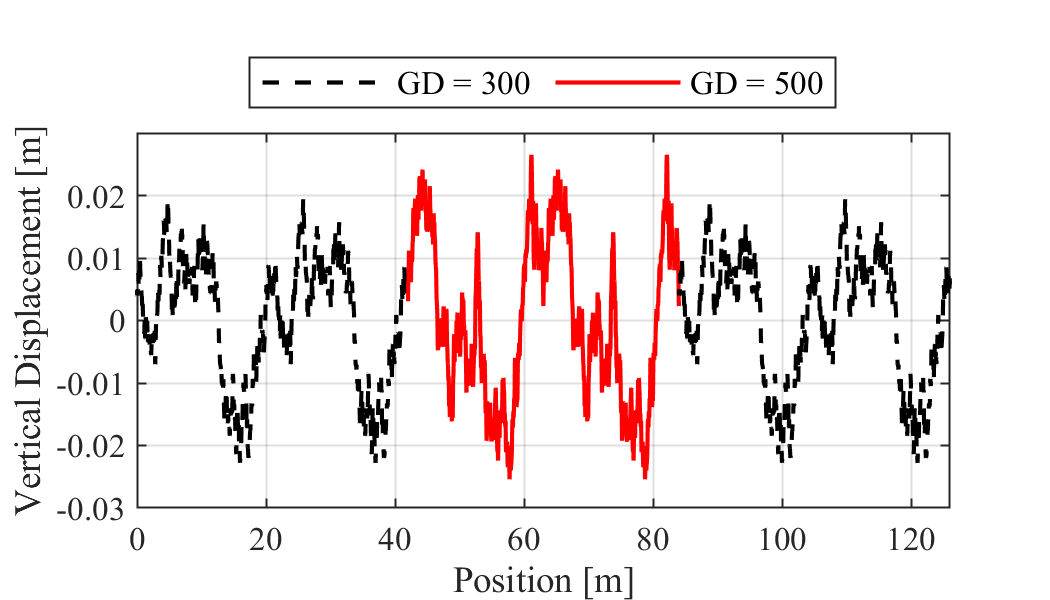}
    \caption{The vertical displacement of the experimental profile with three segments, each having distinct \texttt{GD} values depicted by different line types and colors.}
    \label{fig3}
\end{figure}

To test the efficacy of our approach, a 150 m long test surface was created with varying \texttt{GD} values, segmented into 3 parts: the first and last 50 m with a prescribed \texttt{GD} of 300, and the second 50 m with a prescribed \texttt{GD} of 500 as shown in Fig \ref{fig:simulation}. These surfaces were created based off of segments of originally 21 m long surfaces that were created at the intended \texttt{GD} values. Figure \ref{fig3} illustrates the vertical displacement profile of these segments with distinct \texttt{GD} values. During the simulations, the model controls the robot along with collecting IMU data from the sensors to provide as feedback to the BCR. The IMU data is batched, filtered and processed to transfer only the variance of the longitudinal acceleration values ($\boldsymbol{a_{front}}$) and linear velocity ($v$) to the BCR. The buffer size for the variance of $\boldsymbol{a_{front}}$ is 1000 sample points during run-time at a rate of 120 Hz. There were two different types of test cases trained. Test Case A utilizes the surrogate model trained without noise, and it also ignores noise when collecting data and calibrating the \texttt{GD} value of a specified surface. Test Case B utilizes the surrogate model trained with measurement noise, and it utilizes data with measurement noise when calibrating the \texttt{GD} value of a specified surface. It should be noted that for Test Case B, the distribution of the added measurement noise is described in Eq. \ref{meas_eq}.

\subsection{Results for Test Case A}\label{ResultsA}
\begin{figure}[h]
    \centering
    \includegraphics[scale=0.5, width=90mm]{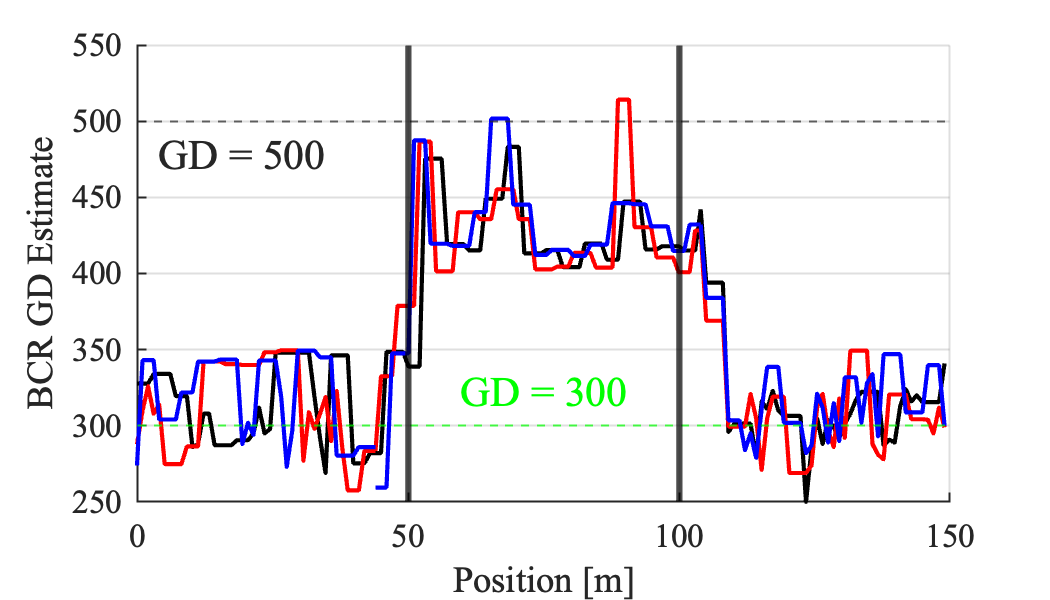}
    \caption{Variation in \texttt{GD} values over the terrain as measured by the BCR for Test Case A. The \textcolor{black}{black} line corresponds to the change in roughness of the terrain shown in Figure \ref{fig:simulation}.}
    \label{fig:GD_plots}
\end{figure}

\begin{figure}[h]
    \centering
    \includegraphics[scale=0.5, width=90mm]{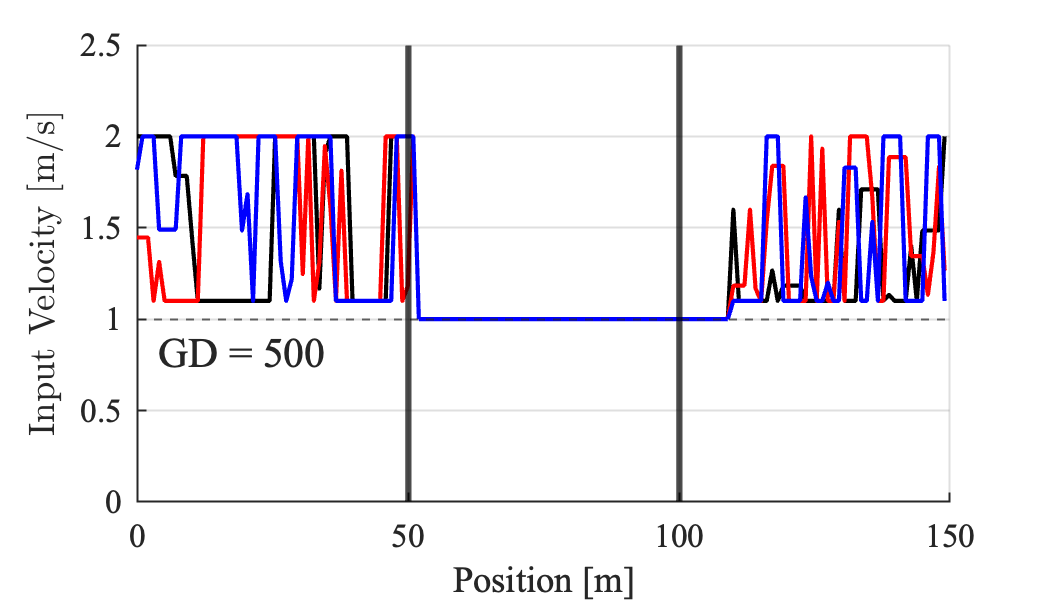}
    \caption{Variation in linear velocity input $(v)$ over the terrain, modulated by the Simplex controller for Test Case A. The \textcolor{black}{black} line corresponds to the change in terrain roughness shown in Figure \ref{fig:simulation}.}
    \label{fig:vel_plots}
\end{figure}

Test Case A refers to the situation where the surrogate model is trained without measurement noise, and the data collected for the surface's \texttt{GD} calibration assumes an absence of measurement noise as well. 
Once the BCR was trained, simulations were conducted on the profile specified in Figure \ref{fig3}. The range of GD values in the performance mode is set as $[\texttt{GD}_{min},\texttt{GD}_{max}]$ = [250,350]. We set the maximum allowable velocity to 2.0 m/s for the Jackal robot on the performance mode terrains. When the BCR estimates $\texttt{GD} > 350$, the Simplex controller switches from \enquote*{Performance} to \enquote*{Safety} mode, lowering the velocity to a constant 1.0 m/s. The control algorithm follows its mathematical definition as stated in Equation \ref{control_eq}. Figures \ref{fig:GD_plots} and \ref{fig:vel_plots} show the variation in estimated \texttt{GD} values and the input linear velocity as the robot traverses the terrain. It should be noted that this test case was run 3 times to help show how the calibration is stochastic and
will vary between calibrations over the same test surface. Around the 50 m mark, the BCR detects a terrain roughness increase, and the Simplex controller adjusts accordingly. This confirms the effectiveness of our method in real-time adaptation. Additionally, at around the 100 m mark, the BCR detects the terrain roughness decrease, and the Simplex controller adjusts once again. 

To quantify the amount of error in this calibration process, the root mean squared error (RMSE) technique was used. The RMSE was considered over regions where the prescribed \texttt{GD} was 300 and regions where the prescribed \texttt{GD} was 500. For Test Case A, the RMSE over regions with a prescribed \texttt{GD} of 300 was found to be 42.58, and the RMSE over regions with a prescribed \texttt{GD} of 500 was found to be 94.42. This indicates that numerically, the BCR has better calibration at the lower prescribed \texttt{GD} of 300. Visually, this is confirmed in Figure \ref{fig:GD_plots} with the consistent underpredicting of the prescribed \texttt{GD} of 500. There are two probable sources for this behavior: 1) there are a less training points in the surrogate model training at this location, or 2) the stochastic nature of the surface creation leads to a surface not well captured by the training data. To help showcase the stochastic nature of the BCR, the calibration was done over the same surface 3 times.

\subsection{Results for Test Case B}\label{ResultsB}
\begin{figure}[h]
    \centering
    \includegraphics[scale=0.5, width=90mm]{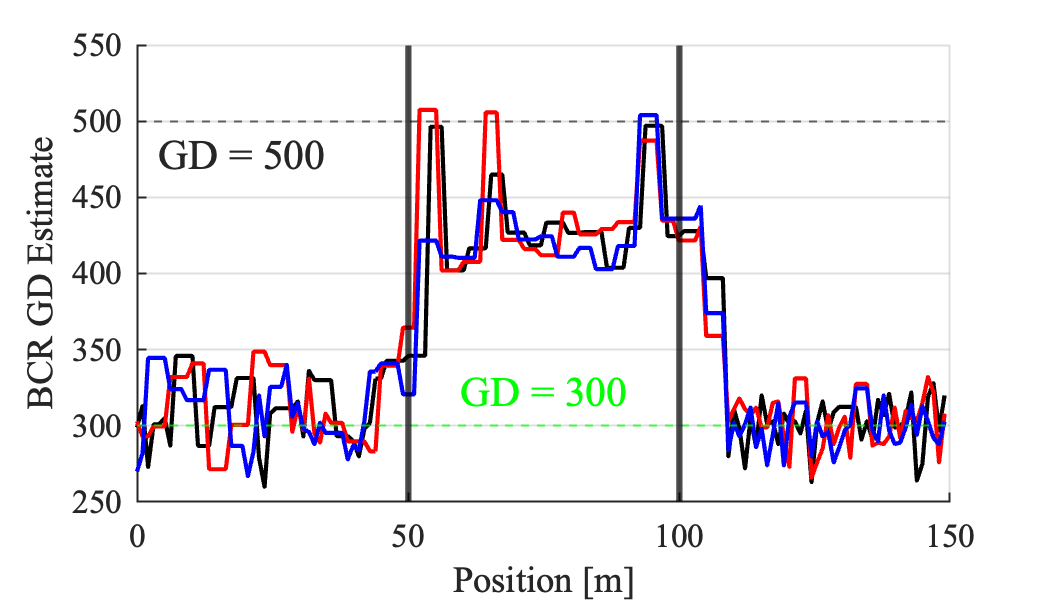}
    \caption{Variation in \texttt{GD} values over the terrain as measured by the BCR for Test Case B. The \textcolor{black}{black} line corresponds to the change in roughness of the terrain shown in Figure \ref{fig:simulation}.}
    \label{fig:GD_plots_B}
\end{figure}

\begin{figure}[h]
    \centering
    \includegraphics[scale=0.5, width=90mm]{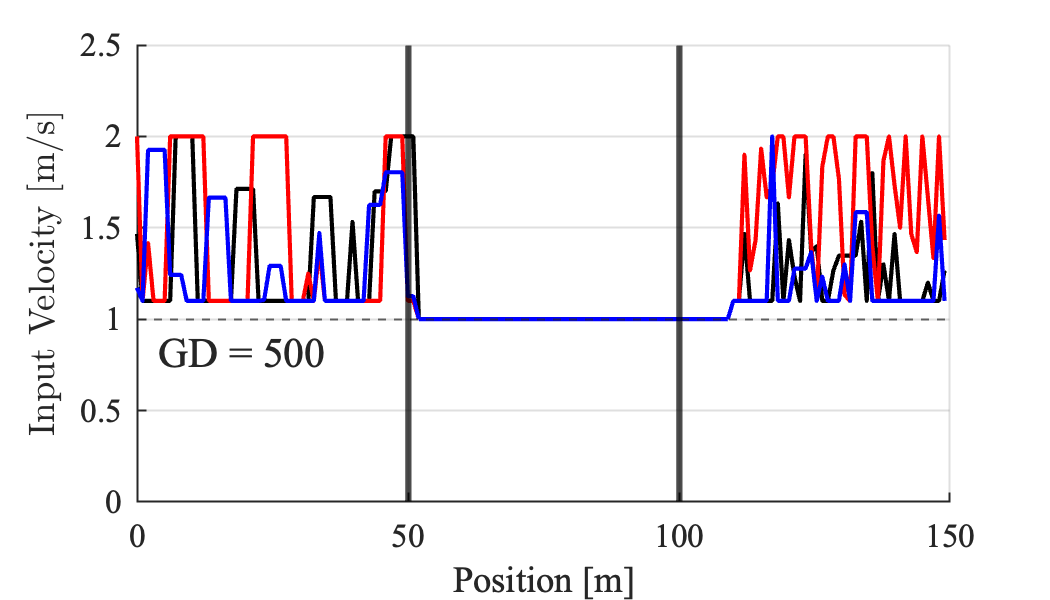}
    \caption{Variation in linear velocity input $(v)$ over the terrain, modulated by the Simplex controller for Test Case B. The \textcolor{black}{black} line corresponds to the change in terrain roughness shown in Figure \ref{fig:simulation}.}
    \label{fig:vel_plots_B}
\end{figure}

Test Case B refers to the situation where the surrogate model is trained with measurement noise, and the data collected for the surface's \texttt{GD} calibration assumes the data collection has measurement noise as well. 
Once the BCR was trained, simulations were conducted on the profile specified in Figure \ref{fig3}. And, the control approach used for Test Case A is also used for Test Case B. Figures \ref{fig:GD_plots_B} and \ref{fig:vel_plots_B} show the variation in estimated \texttt{GD} values and the input linear velocity as the robot traverses the terrain. It should be noted that this test case was also run 3 times to help show how the calibration is stochastic and will vary between calibrations over the same test surface. Around the 50 m mark, the BCR detects a terrain roughness increase, and the Simplex controller adjusts accordingly. This confirms the effectiveness of our method in real-time adaptation. Additionally, at around the 100 m mark, the BCR detects the terrain roughness decrease, and the Simplex controller adjusts once again. 

Similarly to Test Case A, the RMSE was used to quantify the error for regions where the prescribed \texttt{GD} was 300 and regions where the prescribed \texttt{GD} was 500. For Test Case B, the RMSE over regions with a prescribed \texttt{GD} of 300 was found to be 50.19, and the RMSE over regions with a prescribed \texttt{GD} of 500 was found to be 96.23. This indicates that Test Case B, like Test Case A, has better calibration at the lower prescribed \texttt{GD} of 300. Visually, this is confirmed in Figure \ref{fig:GD_plots_B} with similar reasonings for the underprediction of that middle surface as Test Case A. It can also be seen here, that Test Case B has higher RMSE values than Test Case A which is expected because Test Case B has the addition of simulated measurement noise in both its training data and its data gathered for calibration.

\section{Conclusion \& Future Work}\label{Conclusion}
Our approach demonstrated effective real-time adaptation to varying terrain roughness, maintaining the safety and performance of a mobile robot.We have successfully showcased the effectiveness of a Simplex controller architecture in switching the mode of operation. The switch between safety and performance modes depends on the estimated \texttt{GD} value of the terrain, which is determined during run-time by the Bayesian Calibration Routine (BCR). Results shown in Figures \ref{fig:vel_plots} and \ref{fig:vel_plots_B} depict the change in control velocities when the robot detects a change in \texttt{GD} value of the surface. It was shown here that the robots were able to successfully leave the \enquote{Safety} mode when the calibrated \texttt{GD} was below a certain threshold. 

The addition of measurement noise in training the surrogate model and calibrating the surface's roughness has added more error to the calibration process. This is reflected in the RMSE values of the calibration. The RMSE values of predicting the surface with the prescribed \texttt{GD} of 300 increased by 18\% when measurement noise was considered, and the RMSE values of predicting the surface with the prescribed \texttt{GD} of 500 was increased by 2\% when measurement noise was considered. This is expected because considering measurement noise adds a level of error to the calibration. The added error level when the prescribed \texttt{GD} is 300 is a bit higher than expected, but it should be noted that it does have an overall lower RMSE value in both cases when compared to the surfaces with the prescribed \texttt{GD} value of 500. 

When determining the calibrated \texttt{GD} of a surface, the Bayesian calibration method doesn't perform as well in situations near the boundaries of the domain. Future work will focus on refining the BCR for more accurate roughness estimation and considering weighted approaches for different velocities to address inaccuracies during calibration. Another direction for future work would be to simulate more complex terrains with a higher level of fidelity for the vehicle model.

\bibliographystyle{ieeetr} 
\bibliography{references}

\section{Contact Information}
Edwina Lewis\\
 Graduate Research Assistant, School of Civil and Environmental Engineering and Earth Sciences, Clemson University \\ 
 e-mail: edwinal@clemson.edu

Aditya Parameshwaran\\
 Graduate Research Assistant, Department of Mechanical Engineering, Clemson University \\ 
 e-mail: aparame@clemson.edu
 
Laura Redmond\\
 Assistant Professor, School of Civil and Environmental Engineering and Earth Sciences and Department of Mechanical Engineering, Clemson University \\ 
 e-mail: lmredmo@clemson.edu
 
 Yue Wang\\
 Professor, Department of Mechanical Engineering, Clemson University \\ 
 e-mail: yue6@g.clemson.edu

\section{Acknowledgements}
This work was supported by Clemson University's Virtual Prototyping of Autonomy Enabled Ground Systems (VIPR-GS), under Cooperative Agreement W56HZV-21-2-0001 with the US Army DEVCOM Ground Vehicle Systems Center (GVSC). Additional thanks is given to Stephen Wright for the work on migrating the GPM/SA code from \texttt{Matlab} to \texttt{Julia}. \\DISTRIBUTION STATEMENT A. Approved for public release; distribution is unlimited. OPSEC9325 

\onecolumn
\section{APPENDIX A}\label{appendix_A}

\begin{figure*}[h]
    \centering
    \includegraphics[scale = 1.0,width = 180mm]{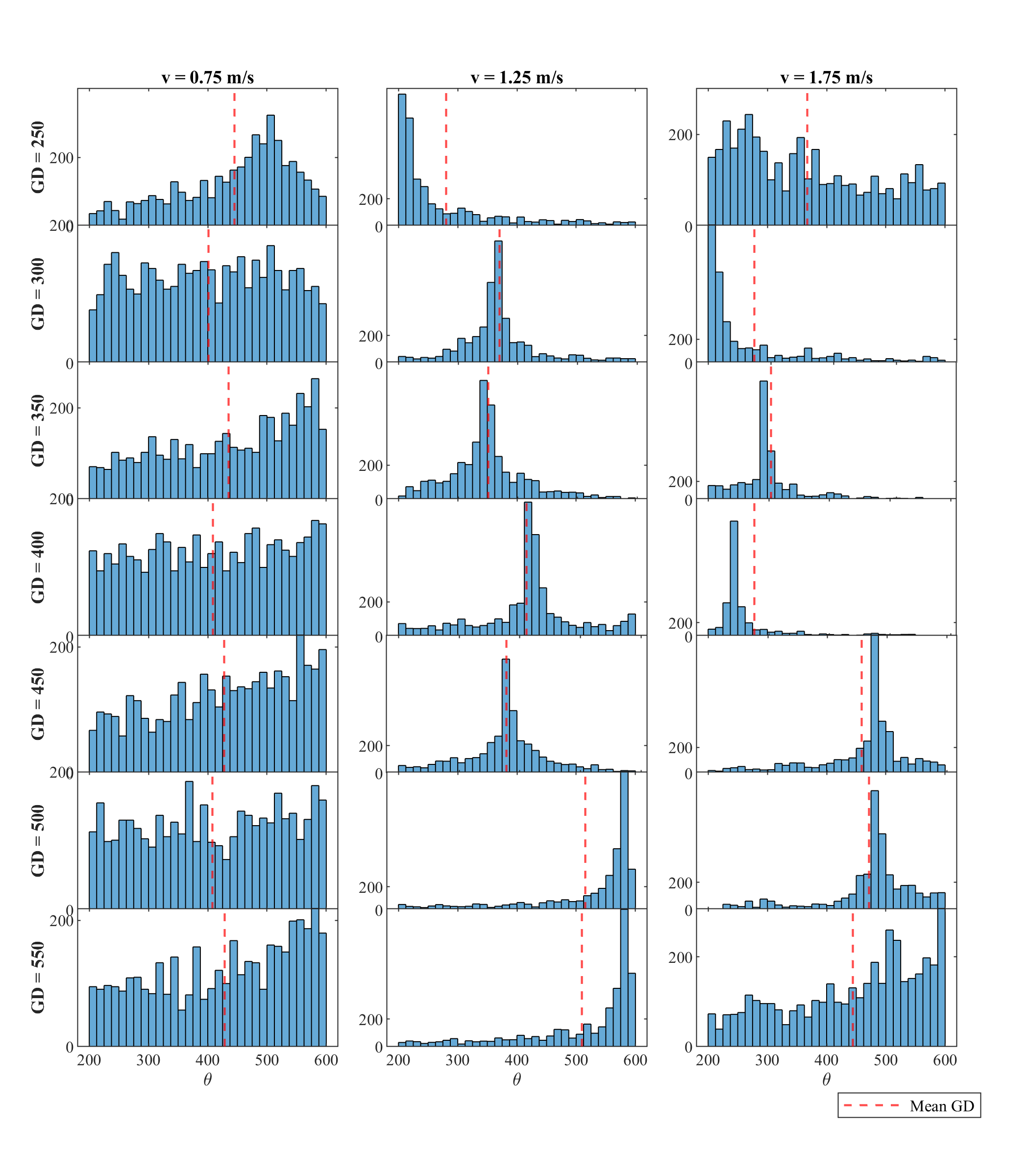}
    \caption{The histograms for various prescribed \texttt{GD} values at differing velocities.}
    \label{fig4}
\end{figure*}

To fully understand whether the surrogate model could act as a good fit for the calibration routine, the posterior distribution of the gathered $\theta$ values had to be assessed. Fig. \ref{fig4} shows the posterior $\theta$ distributions for vehicles traversing surfaces with prescribed \texttt{GD} values of 250, 300, 350, 400, 450, 500, and 550. It should be noted that the surrogate model was trained without measurement noise considered, and the \texttt{GD} value of the surface was calibrated without measurement noise considered.
\clearpage

\end{document}